\documentclass[11pt]{article}

\usepackage[14pt]{extsizes}
\usepackage[font=small]{caption}

\usepackage{parskip}          
\usepackage{times}            
\usepackage{geometry}
\usepackage{graphicx}         
\usepackage{hhline}           
\usepackage{amsfonts}         
\usepackage[rightcaption]{sidecap} 
\usepackage{hyperref}         
\usepackage{listings}         

\usepackage{subfigure}
\usepackage{booktabs}

\usepackage{fancyhdr}
\begin{document}

\title{\vspace*{-3cm}
A Multiple-Choice Test Recognition System\\
based on the Gamera Framework\footnote{Published in C.~Dalitz (Ed.):
``Document Image Analysis with the Gamera Framework.'' Schriftenreihe des
Fachbereichs Elektrotechnik und Informatik, Hochschule Niederrhein, vol.~8,
pp.5-15, Shaker Verlag (2009)}
}
\author{
Andrea Spadaccini\footnote{\texttt{a.spadaccini(at)diit.unict.it}}\\
Dipartimento di Ingegneria Informatica e delle Telecomunicazioni\\
University of Catania\\
Via A. Doria, 5 - Catania (CT) - Italy\\
\and
Vanni G. Rizzo\footnote{\texttt{ascoltalatuasete(at)gmail.com}}\\
Junior Enterprise Catania - JECT\\
Via Sgroppillo, 7 - S. Gregorio (CT) - Italy\\
}
\date{}
\maketitle

\begin{abstract}
\noindent
This article describes JECT-OMR, a system that analyzes digital images
representing scans of multiple-choice tests compiled by students. The system performs a
structural analysis of the document in order to get the chosen answer for each
question, and it also contains a bar-code decoder, used for the identification of
additional information encoded in the document.
JECT-OMR was implemented using the Python programming language, and leverages
the power of the Gamera framework in order to accomplish its task.
The system exhibits an accuracy of over 99\% in the recognition of marked and
non-marked squares representing answers, thus making it suitable for real world applications.
\end{abstract}

\renewcommand{\labelenumi}{\alph{enumi})} 


\section{Introduction}
\label{sec:intro}
Multiple-choice tests are a common way of testing someone's knowledge, and in
Italy they are usually employed, among the many possible fields of application,
during the admission tests for \textit{numerus clausus} University courses like
Medicine \cite{miur}. This led to the foundation of many small enterprises that have the
sole business of training students for getting a good rank by compiling those
tests in the best possible way.

In order to cope with the necessity of automate the correction process for
large numbers of hand-compiled multiple-choice tests, the Junior Enterprise of
Catania (JECT) \cite{ject} - the non profit students association to whom the two authors
belong to - developed JECT-OMR, the system described in this article.

Optical Mark Recognition (OMR) is a process for the detection and recognition of
marks written on paper, that is often employed for the recognition of the
answers checked in multiple-choice tests \cite{lowcost}
\cite{shared-questionnaire}.

The JECT-OMR system takes as input a digital image containing the scan of a paper test
compiled by a student, and returns as output a vector containing, for each
question, the answer that the student selected by tracing a cross in the chosen
square, the blank squares and (if any) symbols with special meanings (like the
ones described in Section~\ref{subsec:cancel}). 

Moreover, the system also returns the values of the barcodes present in
the test, so that the it can be contextualized without manual intervention
simply by interacting with the existing databases.

The programming language used for the implementation of JECT-OMR is Python
\cite{python}, and the heart of the system is the Gamera framework for documents
analysis and recognition \cite{gamera}, that proved to be a valuable tool in
various image analysis contexts \cite{gamera-ocr} \cite{gamera-tpami} and gave
us a nice set of building blocks for our system.

The article is structured as following. The structure of the tests processed by
the system is explained in Section~\ref{sec:test-structure}. Section~\ref{sec:diagram}
describes the system by analyzing its components and
Section~\ref{sec:performance} presents the results of an accuracy test.

\section{Document structure}
\label{sec:test-structure}
There are two kinds of tests that JECT-OMR can analyze, that from now on will be
called \textit{kind A} and \textit{kind B}.

The two models share the number of columns of multiple-choice questions (4) and
the presence of two barcodes.

Here are the main differences between the two kinds of document:
\begin{itemize}
    \item{\textbf{Number of questions per column:} the system expects a document
    composed by four columns of questions and a given number of questions per
    column: the number of questions for each column changes from kind A to kind B;}
    \item{\textbf{Size of the elements in the paper test:} JECT-OMR works with
    fixed-resolution images, and as described in Section~\ref{subsec:segmentation}
    it uses a fixed set of dimensions for each kind of document for the segmentation
    of the image into its different components.}
    \item{\textbf{Ways of canceling answers:} the system lets the user cancel
    already given answers, and, as described in Section~\ref{subsec:cancel}, the
    answers can be canceled in two different ways, one for each kind of test.}
\end{itemize}

\begin{figure}[p]
    \centering
       \includegraphics[width=\textwidth]{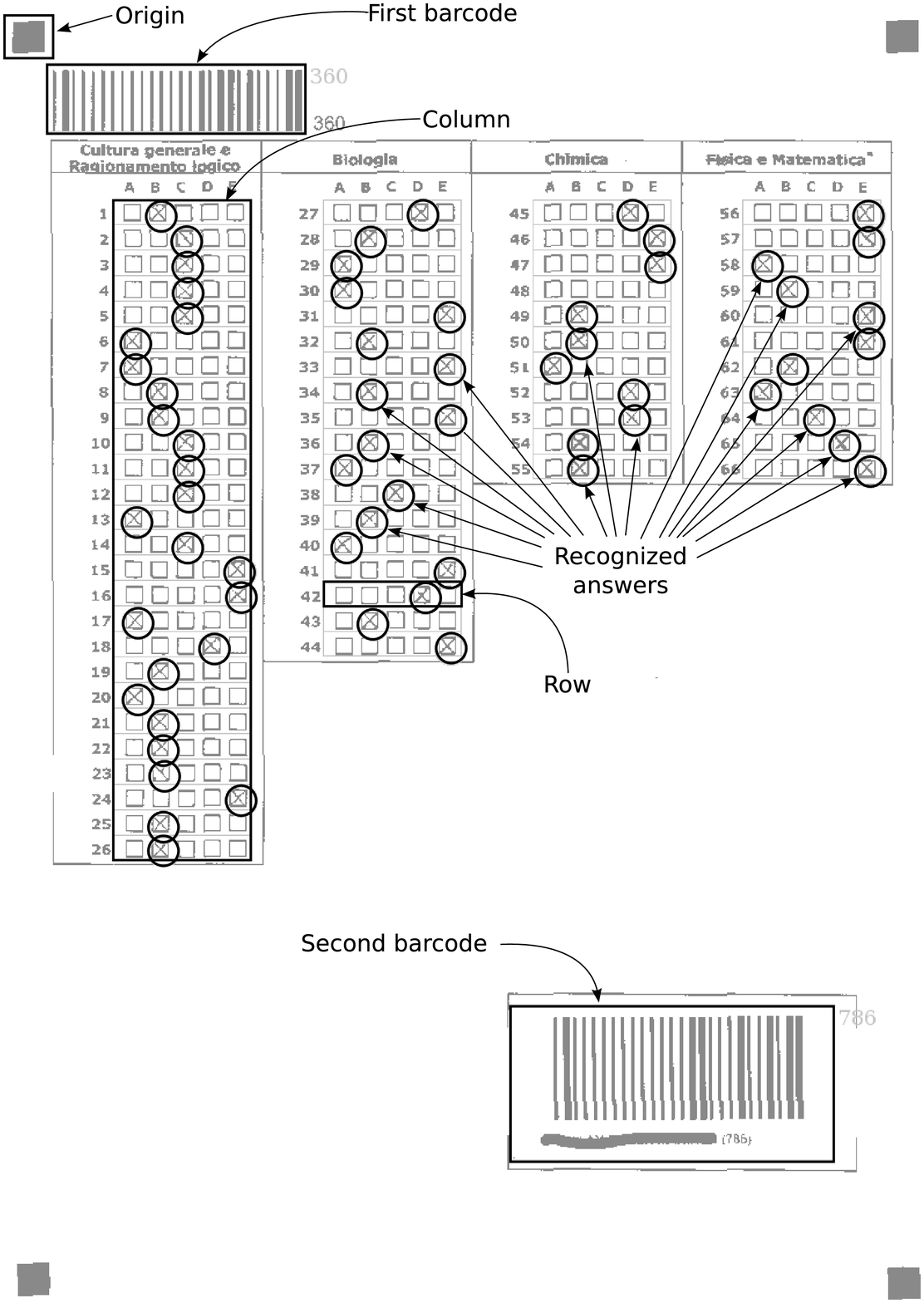}
    \caption{\label{fig:doc-structure} Annotated example document}
\end{figure}

Figure~\ref{fig:doc-structure} shows an annotated example document. The figure
shows in light gray an example document with the highlighted regions found by
the system, and in black there are some annotations that explains the
structure of the document. Of course the annotations point out only some
elements of a given class (e.g.: only few recognized marked answers are pointed
out).

\section{The recognition system}
\begin{figure}[ht]
    \centering
    \includegraphics[width=\textwidth]{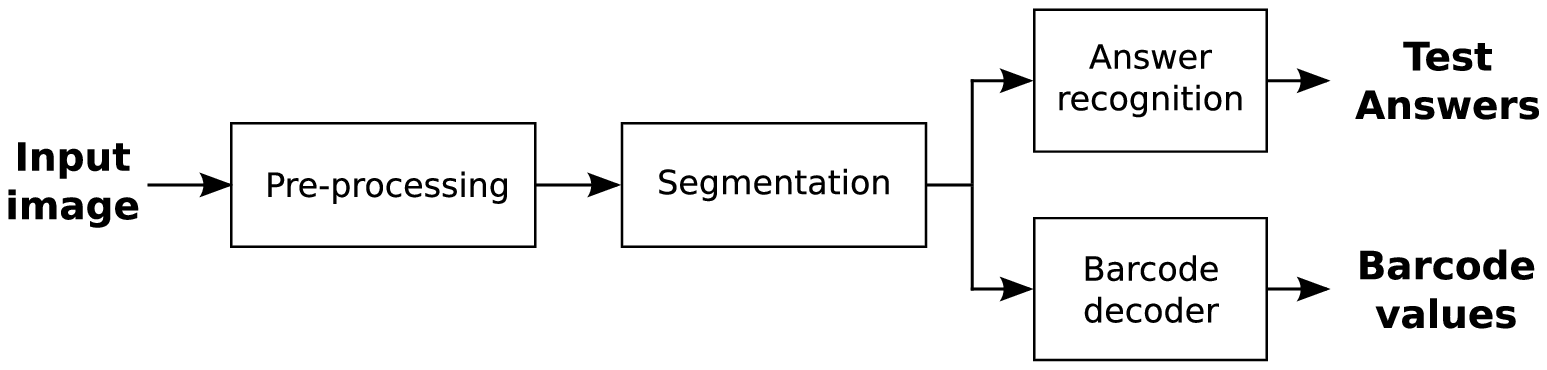}
    \caption{\label{fig:block-diagram} Block diagram of the recognition system}
\end{figure}

In this section the system will be described, and its basic blocks 
will be analyzed in detail. Figure~\ref{fig:block-diagram} shows the block
diagram of the system.
\label{sec:diagram}

\subsection{Pre-processing}
\label{subsec:preprocessing}
The pre-processing phase consists in a set of operations that make the scanned
image more suitable for the further phases.

The first operation performed to the image is the conversion to gray scale; then
the image is converted into black and white format using the global Otsu
thresholding method \cite{otsu}, which is  Gamera's default binarization method.

Next the system does a compensation of rotation effects induced by the scanning
operation. The rotation is estimated and then it is corrected by rotating the
image in the opposite direction by the same angle. The rotation correction is
performed using a projection profile based algorithm built into Gamera\footnote{A detailed description of the skew-detection algorithm can be found in section 3.1 of Ref.~\cite{gamera-rot}}.

In this phase it is clear that using the Gamera framework is a good choice,
because it offers abstractions that allow the programmer to concentrate on the
problem domain. For example, the operations described above are done by 
as little code as shown in Listing~\ref{lst:pre-processing}.

\lstset{language=Python,
  basicstyle=\small \ttfamily,
  frame=bottomline,
  floatplacement=t!,
  aboveskip=0pt,
  captionpos=b
}
\begin{lstlisting}[float, caption=Code of the pre-processing phase,
	label=lst:pre-processing]
    image = image.to_greyscale()
    image = image.otsu_threshold()
    angle = image.rotation_angle_projections(-15, 15)[0]
    image = image.rotate(angle, 0)
\end{lstlisting}

Finally, the system finds the upper left black square, marked as \textit{origin}
in Figure~\ref{fig:doc-structure}, by performing a connected
component analysis and finding the glyph whose coordinates (\texttt{offset\_x}
and \texttt{offset\_y}) are the nearest to the point $(0,0)$.  Those
coordinates are called the \textit{origin} of the test, and are used as the
reference point for any subsequent elaboration.

\subsection{Segmentation}
\label{subsec:segmentation}
After the pre-processing phase, JECT-OMR extracts from the image the regions
containing the barcodes and the compiled questions, using empirically determined
dimensions relative to the origin of the test. The dimensions are expressed in
pixels as the scanning resolution of the input images is fixed.

Figure~\ref{fig:doc-structure} shows some of the regions identified in this
phase.

The four regions that contain the questions are the four columns that will be
processed by the algorithm described in Section~\ref{subsec:answer-recognition}.

As explained in section~\ref{sec:test-structure}, the two kinds of test are
structurally slightly different. This means that the program uses two different
sets of dimensions, one for each kind of test.

\subsection{Answer recognition}
\label{subsec:answer-recognition}
Each column of questions is split into equal-height rows, that are analyzed one by
one.

\begin{figure}[ht]
    \centering
    \includegraphics[width=.5\textwidth]{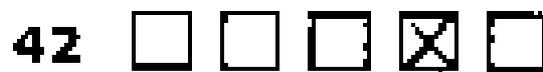}
    \caption{\label{fig:question-42} Example row}
\end{figure}

Each row is an image region that represents a question, from which the chosen
answer (if any) is extracted. Figure~\ref{fig:question-42} shows an example row.

Here is how the answer classification algorithm works. First of all, the image is
split into connected components, that are sorted according to horizontal
position, from left to right. If there are no errors, the system finds 5
components relative to the 5 squares available for choosing the answer, of which
one might be checked.

If the analysis returns a number of component different than 5, the analysis is
revised and repeated: using an adaptive algorithm, that changes the thresholds until it does not find
an acceptable number of components, the system classifies each glyph using two
features: the number of black pixels (\texttt{black\_area}, natively available
in Gamera) and the squareness (computed as \texttt{abs(num\_rows -
num\_columns)}).

Once the glyphs corresponding to squares are found, they are analyzed in order
to classify them as \textit{chosen} squares or \textit{empty} squares. 
Gamera provides a statistical
kNN classifier, but for this specific problem a simple
heuristic approach based on threshold values was sufficient.
The features used in this classification are the number of black pixels and the
number of holes inside the image.

If a square was marked by the student, and thus must be classified as
\textit{chosen}, the number of holes (white regions surrounded by black regions)
will be bigger than if the square were not marked. This can be clearly seen in
Figure~\ref{fig:question-42}, where the fourth square would labeled by the system
as \textit{chosen} and the other four as \textit{empty}.

The condition used to classify each square was found empirically, and it is
composed by 3 sub-conditions:
\begin{enumerate}
    \item{threshold on holeness}
    \item{threshold on black\_area}
    \item{combined threshold over the weighted sum of holeness and black\_area}
\end{enumerate}

Using this algorithm, the system analyzes each row and extracts a 5-elements
vector where the $i$-th position in 1 if the corresponding square is
\textit{chosen}, and 0 if it is \textit{empty}.

The output of this phase is a $n \cdot 5$ matrix, where $n$ is the number of
questions in the test.

\subsubsection{Canceling answers}
\label{subsec:cancel}
When the student is compiling the test, he can undo his actions by canceling his
answers in the following ways: if the test is of kind A, the student can
cancel an answer by making the selected square completely black; if the test is
of kind B, the student can mark the circle that is drawn to the left of the
question, and the system will not consider the answer that the student selected.

The circle used in the kind B is simply the leftmost glyph in the row, and it is 
recognized as full or empty using the \texttt{black\_area} feature.

\begin{figure}[ht]
    \centering
    \includegraphics[width=.5\textwidth]{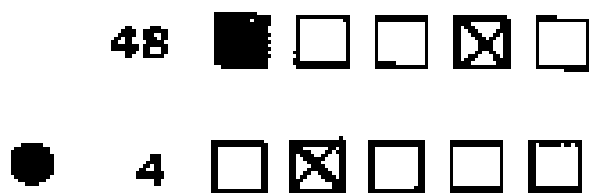}
    \caption{\label{fig:canceled} Examples of canceled questions (top: kind A;
bottom: kind B)}
\end{figure}

Figure~\ref{fig:canceled} shows an example of each of the two cancellation methods.

\subsection{Barcode recognition}
\label{subsec:barcode}
The barcodes recognized by the system are discrete two-width barcodes that
encode binary strings. The information is encoded only in the bars and not in
the spaces, differently from many higher-density barcodes like Interleaved 2 of
5 and Code 128.

The binary string is 26 bits wide, and contains 2 start bits (the first two,
both holding the value 1) and two end bits (holding respectively the values 1
and 0). The bits from 3 to 22 contain the binary representation of the number
encoded in the barcode. The bits 23 and 24 contain the parity of the number of
ones and zeros used for the representation of the number.

This format was used because the software that prints the paper tests recognized
by JECT-OMR uses it, so the system must be able to understand it.

The rest of this section describes how the barcode recognition algorithm works.

Each region containing a barcode is first processed by the rotation correction
algorithm described in Section~\ref{subsec:preprocessing}, because the
recognition algorithm requires that the barcode lines are vertical. The
algorithm is executed for each barcode because some of them are not printed
directly into the test, but are applied by the student using a sticker. This
means that usually they will not be parallel to the rest of the test, and that
they could need another adjustment in order to be suitable for the recognition.

Next, the region is vertically split in 5 sub-regions, each of them is analyzed
by the same recognition algorithm.

This algorithm does as its first step a connected component analysis, and sorts
the resulting glyphs by horizontal position.

Next the algorithm measures the $x$-extension of each glyph. If its value is higher
than a given threshold, the glyph represents the value 1, otherwise it
represents a 0.

From each sub-region the algorithm computes a number in binary form, that is
then converted to decimal and compared to the values obtained from the other
sub-regions.

The algorithm returns the number that is obtained from the highest number of
sub-regions. This implies that if the barcode is slightly damaged it can be
read regardless of the damage.

\section{Error handling}
\label{subsec:correction}
Sometimes it happens that the system cannot perform its tasks because of some
unknown error, maybe because of low-quality scans or unexpected pen strokes made
by the students.

When error conditions, like more or less than five squares in a row,
are detected by the system, it adds the number of the
question being analyzed to a list of rows that could not be processed, and shows
to the user a simple Graphical User Interface (GUI) for error correction.

The GUI was written using wxPython \cite{wxpython}, the Python bindings of the cross-platform
user interface toolkit wxWidgets, and shows to the user an image of the scanned
test where the results of the elaboration are marked using different colors for
the different regions of the image, and lets him correct manually the recognition
errors detected earlier by the system.

\section{Performance evaluation}
\label{sec:performance}
In order to evaluate the performance of the system, we measured its accuracy in
two crucial tasks: the recognition of squares related to answers in
multiple-choice questions and the recognition of barcodes.

The results of the accuracy tests are shown in Table~\ref{tab:results}.

\begin{table}[hbp]
  \begin{center}
    \begin{tabular}{lccc}
      \toprule
      Class & Number of samples & Accuracy & Error\\
      \midrule
      Barcode & 190 & 98 \% & 2 \% \\
      Marked square & 38300 & 99.1 \% & 0.09 \% \\
      Non-marked square & 186857 & 100 \% & 0 \% \\
      \bottomrule
    \end{tabular}
  \end{center}
  \caption{Results of the performance evaluation}
  \label{tab:results}
\end{table}

\section{Conclusion}
\label{sec:concl}
This work presented JECT-OMR, a recognition system for multiple-choice tests
based on the Gamera framework.

Using the Gamera framework helped us developing the application in a very short
time, and, due to its intuitive use, we could focus on the application domain
rather than on programming techniques.

The results of the tests show that the application is quite mature, and it has
been used for more than two years by a small enterprise whose mission is to
prepare students for multiple-choice tests used in the admission test of
numerus clausus University courses.

Although the recognition approach based on fixed thresholds proved to work
fairly well, in future we might evaluate a recognition algorithm based on the
kNN classifier built into Gamera, as it would probably be easier to use and more
robust to unforeseen usage conditions or marks.

Any other future work will probably be done for improving the usability of the error
correction GUI and for the implementation of new features as the Italian
Education Ministry changes the mechanism of the test for numerus clausus courses.



\begin{thebibliography}{9}
    \raggedright

    \bibitem{miur} Italian Ministry of Education: {\em Accesso Programmato.}
    \url{http://accessoprogrammato.miur.it} 

    \bibitem{ject} Junior Enterprise Catania: {\em JECT Homepage.}
    \url{http://www.ject.it} 

    \bibitem{lowcost} H.~Deng, F.~Wang, B.~Liang: {\em A Low-Cost OMR Solution
    for Educational Applications.} International Symposium on Parallel and
    Distributed Processing with Applications, pp.~967-970 (2008)

    \bibitem{shared-questionnaire} H.~Kubo, H,~Ohashi, M.~Tamamura, T.~Kowata,
    I.~Kaneko: {\em Shared Questionnaire System for School Community Management.}
    International Symposium on Applications and the Internet Workshops,
    pp.~439-445 (2004)

    \bibitem{python} G.~van~Rossum: {\em Python tutorial.} Technical Report CS-R9526,
    Centrum voor Wiskunde en Informatica (CWI), Amsterdam, May 1995.\\
    (see also \url{http://python.org/})

    \bibitem{gamera} M.~Droettboom, C.~Dalitz: {\em The Gamera Homepage.}
    \url{http://gamera.informatik.hsnr.de/} (2008-)

    \bibitem{gamera-ocr} S.~Reddy, G.~Crane: {\em  A Document Recognition System
    for Early Modern Latin.} Chicago Colloquium on Digital Humanities and Computer
    Science (2006)

    \bibitem{gamera-tpami} C.~Dalitz, M.~Droettboom, B.~Pranzas: {\em  A
    Comparative Study of Staff Removal Algorithms.} IEEE Transactions on Pattern
    Analysis and Machine Intelligence 30, pp.~753-766 (2008)

    \bibitem{otsu} N.~Otsu: {\em A threshold selection method from gray-level
    histograms.} IEEE Transactions on Systems, Man, and Cybernetics 9,
    pp.~62-66 (1979)

    
    \bibitem{gamera-rot} C.~Dalitz, G.K.~Michalakis, C.~Pranzas: {\em Optical Recognition of
    Psaltic Byzantine Chant Notation.} International Journal of Document
    Analysis and Recognition 11, pp.~143-158 (2008)

    \bibitem{wxpython} R.~Dunn, N.~Rappin: {\em wxPython in Action.}
    Manning. (2006)\\
    (see also \url{http://wxpython.org/})

\end{thebibliography}
\end{document}